\documentclass[submission,copyright,creativecommons]{eptcs}
\providecommand{\event}{ICLP2020, Special Session: Women in Logic Programming} % Name of the event you are submitting to
\usepackage{underscore}           % Only needed if you use pdflatex.
\usepackage{times}
\usepackage{graphicx}
\usepackage{latexsym}
\usepackage{tikz}
\usetikzlibrary{arrows,automata,positioning}
\usepackage[all]{xy}
\usepackage{xspace}
\usepackage{xurl}
\usepackage{listings}
\usepackage{fancyvrb}
\usepackage{xcolor}
\usepackage{tabularx}

\newcommand{\vmnote}[1]{}

\title{Logical Judges Challenge Human Judges\\ on the Strange Case of B.C.--Valjean}
\author{Viviana Mascardi
\institute{University of Genova, DIBRIS, Italy}
\email{viviana.mascardi@unige.it}
\and
Domenico Pellegrini
\institute{Ministry of Justice, Tribunale di Genova, Italy}
\email{domenico.pellegrini@giustizia.it}
}

\def\titlerunning{Logical Judges Challenge Human Judges on the Strange Case of Jean Valjean}
\def\authorrunning{V. Mascardi  \& D. Pellegrini}
\begin{document}
\maketitle

\begin{abstract}
On May 12th, 2020, during the course entitled {\em Artificial Intelligence and Jurisdiction Practice} organized by the Italian School of Magistracy, more than 70 magistrates followed our demonstration of a Prolog logical judge reasoning on an armed robbery case. Although the implemented logical judge is just an exercise of knowledge representation and simple deductive reasoning, a practical demonstration of an automated reasoning tool to such a large audience of potential end-users represents a first and unique attempt in Italy and, to the best of our knowledge, in the international panorama. In this paper we present the case addressed by the logical judge -- a real case already addressed by a human judge in 2015 -- and the feedback on the demonstration collected from the attendees. 
 %The working group involved more than 70 attendees. We translated the evidences of a real armed robbery case into Prolog and we implemented a prototypical, iper-simplified ``logical judge''. 
%After an introductory presentation of artificial intelligence and logic programming, we run a demo to show the conclusions reached by the logical judge based on the available evidences.
%, we drew a parallel between some assumptions made in logic programming and in the Italian law, and we engaged into a lively discussion with the attendees.
%The case study was used to disseminate the potential of AI at the Italian "Scuola Superiore della Magistratura" in a course held on May 2020.
\end{abstract}

\section{Introduction}
\label{section:introduction}

The connections between logic programming and law have been studied for a  long time. In 1975, Meldman discussed his PhD Thesis entitled ``A preliminary study in computer-aided legal analysis'' \cite{meldman1975preliminary} where he modelled legal facts in a Lisp-like language and used instantiation (recalling unification) and syllogism (recalling resolution) to perform a simple kind of
legal analysis inspired by Prosser's Law of Torts \cite{prosser1941handbook}. At that time Prolog was just born, but its applications to legal reasoning were not long in coming. One of the first attempts was made by Hustler \cite{hustler1982programming} who implemented a prototype of a legal consultant in Prolog, again inspired by Prosser's work. A few years later, Kowalski, Sergot et al. succeeded  in   running   a   significant   portion   of   the   1981   British  Nationality  Act,   implemented  in   Prolog  on  a  small  micro  computer 
%\cite{DBLP:conf/ijcai/KowalskiS85,DBLP:journals/cacm/SergotSKKHC86}. 
\cite{DBLP:journals/cacm/SergotSKKHC86}. 
In the same years, Prolog became very popular for implementing expert systems for the legal domain %\cite{10.5555/91456.91463,mackaay1990logic,yoshino1986legal}.
\cite{10.5555/91456.91463,yoshino1986legal}.

From those early attempts, much progress has been made: research on deontic and defeasible reasoning %\cite{benzmuller2020designing,DBLP:conf/jurix/CalegariCLOS19,lam2019enabling}, 
\cite{benzmuller2020designing,DBLP:conf/jurix/CalegariCLOS19}, 
ontological reasoning 
%\cite{el2017towards,wyner2012legal}, 
\cite{el2017towards}, 
and argumentation \cite{governatori2018deontic,walton2018legal} is extremely lively and helps disclosing the many connections between logic programming (and, more in general, computational logic and automated reasoning) and legal reasoning. The application of automated reasoning to digital forensics is another promising research direction %\cite{costantini2019digital,DFIRE,liu2017identifying} 
\cite{costantini2019digital} 
whose potential is witnessed by the ongoing ``Digital Forensics: Evidence Analysis via Intelligent Systems and Practices'' (DigForASP) COST Action\footnote{CA17124, \url{https://digforasp.uca.es}, funded for four years starting from 09/2018 by the European Cooperation in Science and Technology (COST, \url{www.cost.eu}).}. DigForASP exploits computational logic to reason on crimes evidences to reconstruct possible scenarios related to the crime, even when knowledge is fragmented and incomplete.  

Despite this long and successful history, the potential and limitations of automated reasoning are still obscure to their end-users, namely judges, magistrates, lawyers and prosecutors. These professionals are overwhelmed by news about robotic judges but are not always fully aware of the techniques behind these robotic surrogates; evaluating their pros and cons in an informed and objective way is often out of their reach. 
\vmnote{Nonostante questa storia lunga e di successo, i potenziali e i limiti del ragionamento automatico sono ancora oscuri per i loro utenti finali, vale a dire giudici e magistrati, ma anche avvocati e pubblici ministeri. Questi professionisti sono inondati da notizie sui giudici robotici ma non sono sempre pienamente consapevoli delle tecniche alla base di questi surrogati robotici; valutare i loro pro e contro in modo informato e oggettivo \'e spesso fuori dalla loro portata.
}

At the time being, all the ``robotic judges''
%\footnote{We deliberately use this far from scientific term, as it is very  popular to tag sensational pieces of news related with the exploitation of computer-based support in courts.} 
used in trials employ machine learning, most often deep learning, and many of them became famous for their biased decisions. The  {\em State v Loomis 881 N.W.2d 749 (Wis. 2016) case} is one among the most well known examples: the Wisconsin Supreme Court upheld a lower court's sentencing decision informed by a COMPAS risk assessment report and rejected the defendant's appeal on the grounds of the right to due process.
COMPAS (Correctional Offender Management Profiling for Alternative Sanctions) is a case management and decision support tool developed and owned by Equivant\footnote{\url{https://www.equivant.com/}, last accessed June 2020.}. It is opaque from two points of view: legal, because it is a commercial software whose source code cannot be inspected, and technical, because it employs machine learning techniques which, in many cases including their more recent and successful ``deep'' evolution, are black boxes \cite{10.1093/ijlit/eaz001}.
Using machine learning for boosting predictive justice is becoming a very lively research field, although in many cases the developed applications are academic prototypes, not yet used in real trials. Applications range from predicting decisions of the European Court of Human Rights %\cite{DBLP:journals/peerj-cs/AletrasTPL16,DBLP:journals/ail/MedvedevaVW20} 
\cite{DBLP:journals/ail/MedvedevaVW20}
to predicting recidivism of many different crimes %\cite{butsara2019predicting,ting2018predicting}, 
\cite{butsara2019predicting}, 
to risk assessment in criminal justice \cite{berk2019machine}.

On the one hand, many non technical papers foresee the rise of robotic judges suggesting that they might substitute human judges in most of their activities. On the other, many scientists warn about opaque predictive models also from a technical point of view, besides an ethical one \cite{tolan2019machine}, and advocate the adoption of interpretable models instead \cite{rudin2019stop}. In between, human judges, whose computer literacy is often a basic one, are more and more confused. 
\vmnote{
Da un lato, molti articoli non tecnici prevedono l'ascesa di giudici robotici e suggeriscono che potrebbero sostituire i giudici umani nella maggior parte delle loro attivit\`a. Dall'altro, molti scienziati mettono in guardia dai modelli predittivi opachi anche da un punto di vista tecnico, oltre che da quello etico \cite{tolan2019machine}, e sostengono invece l'adozione di modelli interpretabili \cite{rudin2019stop}. In mezzo, i giudici umani, la cui alfabetizzazione informatica \`e spesso limitata, sono sempre pi\`u confusi.
}

To address this pressing need of clarity, on the 12th and 13th of May 2020 the Italian School of Magistracy (Scuola Superiore della Magistratura, SSM) offered a course entitled {\em Artificial Intelligence and Jurisdiction Practice}. 
%
%The Italian School of Magistracy (Scuola Superiore della Magistratura, SSM) is an autonomous Italian body founded in 2006 with exclusive competence in the training and professional updating of members of the judiciary and magistracy in Italy.
%
%On the 12th and 13th of May 2020, the SSM offered a course entitled {\em Artificial Intelligence and Jurisdiction Practice}. 
%Initially foreseen for mid March 2020 as a standard event, the course was transformed into an online event and shifted of two months due to the Covid-19 pandemics.
%
Within that course, we were in charge for the working group on {\em Criminal Law} carried out during the afternoon of May 12th. The working group was run in parallel with the {\em Civil Law} group and involved more than 70 attendees, one half of the total number of attendees of the course. 
The design of the working group activities required some initial effort due to the mismatches between the vocabulary of the two authors -- a magistrate and a computer scientist -- with a completely different background. Once we aligned our shared terminology, the set up of the activities and the preparation of the teaching material run smoothly and many connections between the Italian law and computational logics were discovered, from the adoption of modus ponens as a well known oratory technique \cite{modusPonens2014} to closed world assumption\footnote{The article 530, second paragraph, of the Italian Code of Criminal Procedure states that ``the judge acquits the defendant also in case {\em there is no evidence of the crime, or the evidence is not sufficient, or it is contradictory}'', which closely resembles Prolog negation as failure and closed world assumption.}.
\vmnote{
Per far fronte a questo urgente bisogno di chiarezza, il 12 e 13 maggio 2020 la Scuola Superiore della Magistratura, SSM, ha offerto un corso intitolato {\em Intelligenza Artificiale e Pratica della Giurisdizione}.
All'interno di quel corso eravamo responsabili del gruppo di lavoro su {\em Diritto Penale} svolto nel primo pomeriggio. Il gruppo di lavoro ha lavorato in parallelo con il gruppo {\em Diritto Civile} e ha coinvolto più di 70 partecipanti, la met\`a dei partecipanti al corso.
La progettazione delle attivit\`a del gruppo di lavoro ha richiesto uno sforzo iniziale a causa delle discrepanze tra il vocabolario dei due autori -- un magistrato e un informatico -- con un bagaglio culturale completamente diverso. Una volta allineata la nostra terminologia condivisa, l'avvio delle attivit\'a e la preparazione del materiale didattico sono andate avanti senza difficolt\`a e si sono scoperti molti collegamenti tra la legge italiana e la logica computazionale, dal modus ponens [........] all'assunzione di mondo chiuso [L'articolo 530, secondo comma, del codice di procedura penale italiano stabilisce che ``...'', che ricorda da vicino la negazione come fallimento finito di Prolog e l'assunzione di mondo chiuso.].
}

To allow the attendees to have an idea of how computational logic might serve their needs, having less than two hours at our disposal, one week before the course took place we sent them an exercise based on a real,  published case of armed robbery, properly obfuscated to avoid that they could recognize it. They had all the details of the case and they were asked to answer some questions based on the availability of evidences and on the reliability of witnesses. In the meanwhile, we translated the robbery facts and evidences into Prolog and we implemented the logical judge. During the working group, after an introductory presentation to artificial intelligence and logic programming, we run a demo to show the conclusions reached by the logical judge based on the available evidences, and we entered a discussion on whether, and how, the logical judge could reach the same conclusions as the human ones. 
%, we drew a parallel between some assumptions made in logic programming and in the Italian law, and we engaged into a lively discussion with the attendees.
\vmnote{
Per consentire ai partecipanti di avere un'idea di come la logica computazionale possa soddisfare i loro bisogni, avendo meno di due ore a nostra disposizione, una settimana prima del corso abbiamo inviato loro un esercizio basato su un caso reale di rapina a mano armata, opportunamente offuscato per evitare che potessero riconoscerlo. I partecipanti avevano tutti i dettagli del caso e dovevano rispondere ad alcune domande in base alla disponibilit\`a delle prove e all'affidabilit\`a dei testimoni. Nel frattempo, abbiamo tradotto i fatti e le prove in Prolog e abbiamo implementato un giudice logico semplificato. Durante le attivit\`a del gruppo di lavoro, dopo un'introduzione all'intelligenza artificiale e alla programmazione logica, abbiamo svolto una dimostrazione per mostrare le conclusioni raggiunte dal giudice logico sulla base delle prove disponibili e abbiamo avviato una discussione su se e come il giudice logico possa raggiungere le stesse conclusioni di quello umano.}

Although our logical judge is no more than a simple exercise of knowledge representation in Prolog, showing what automated reasoning in the legal domain could achieve via a practical demonstration to more than 70 magistrates represents a first and unique attempt in Italy and, to the best of our knowledge, in the international panorama.
\vmnote{Sebbene il nostro giudice logico non sia altro che un semplice esercizio di rappresentazione della conoscenza in Prolog, mostrare ci\`o che il ragionamento automatizzato in ambito legale potrebbe permettere attraverso una dimostrazione pratica a oltre 70 magistrati rappresenta un primo e unico tentativo in Italia e, per quanto ci risulta, nel panorama internazionale.}

%The paper is organized as follows: Section \ref{section:valjeanjavert} describes the armed robbery case, Section \ref{section:logicaljudge} presents its translation into Prolog, Section \ref{section:feedbackattendees} discusses the feedback we got from some attendees via an online, anonymous questionnaire. Section \ref{section:conclusions} concludes and outlines the future directions of our work.

\section{The Human Judge and the Case of B.C.}
\label{section:valjeanjavert}

The armed robbery case we took inspiration from is the case of B.C.\footnote{\texttt{https://archiviodpc.dirittopenaleuomo.org/upload/1445325933Trib\_MI\_Gennari.pdf}, in Italian, last accessed on June 2020.}
To avoid disclosing the case, we named B.C. {\em Valjean} and we used other names from Les Mis\'erables by Victor Hugo for other actors. We also added some further evidences and conditions to make the reasoning more involved\footnote{The full description of the exercise is available, in Italian, here: \url{https://www.dropbox.com/s/8aof95p1q7ajx7k/Esercizio.pdf?dl=0}.} 
The revised B.C. case is the following.
A criminal wearing a red jacket and a full-face motorcycle helmet enters the ABC supermarket wielding a gun, together with his partner in crime. He threatens Enjolras with the gun, asks for the money, and hits him. The two criminals try to get the money but, due to the prompt and unexpected reaction of Enjolras, after a few minutes run away on a scooter. 
During the trial, the following evidences emerge: 

\noindent -- {\bf E1}:
Fantine, a highly reliable witness,  declares that the two criminals  left the supermarket on board of a scooter with plate 12345 at 15.00, more or less. The scooter -- whose theft had been reported a few days before -- was found later in the afternoon.

\noindent-- {\bf E2}: on the scooter's rearview mirror,  the scientific police finds a fingerprint highly compliant with Valjean's one.

\noindent -- {\bf E3}: Fantine also asserts that, before leaving the supermarket, the criminal with the red jacked said to his partner ``Jamunindi, jamunindi!''. This sentence means ``Let's go'' in the dialect from Reggio Calabria, and  Valjean was born in Reggio Calabria.

\noindent -- {\bf E4}: Thenardier asserts that he saw  Valjean at 15.05 riding a scooter with plate 12345, in a road very close to the ABC supermarket, together with another man.

\noindent -- {\bf E5}: in a sound track extracted from a mobile phone accidentally retrieved in the supermarket proximity, dating back 14.55 of the robbery day, a voice that turned out to be Valjean's one can be heard.  

\noindent -- W.r.t. {\bf E4}, the defense lawyer presents evidences that Thenardier is an unreliable witness.  \\

\noindent The exercise we proposed to the attendees is the following:
%we report the solutions based on the Italian Code of Criminal Procedure and on the final outcome of the B.C. case:
based on the declarations of the witnesses, can we demonstrate that the criminal in red jacket is Valjean...

\noindent -- {\bf Q1} ...using evidences {\bf E1, E2, E3}? 

\noindent -- {\bf Q2} ...using evidences {\bf E1, E2, E3, E4}, without using the evidences from the defense lawyer on Thenardier's unreliability? 

\noindent -- {\bf Q3} ...using evidences {\bf E1, E2, E3, E4} and evidences from the defense lawyer? 

\noindent -- {\bf Q4} ...using evidences {\bf E1, E2, E3, E5}? 
\\

\noindent To compute the answer, we must consider that ``the existence of a fact cannot be deduced by evidences, unless they are {\em severe}, {\em precise} and {\em coherent}'' (article 192 of the Italian Code of Criminal Procedure). W.r.t. the fact ``Valjean is the criminal with the red jacket'', 
{\bf E1} alone is not even an evidence, as it does not link the criminal with Valjean;  {\bf E2} is severe but not precise, since there are explanations for Valjean's presence of the fingerprint on the scooter's mirror other than assuming he was riding it on the robbery day; {\bf E3} is neither severe nor precise, since many persons besides Valjean can say ``Jamunindi'' and being born in Reggio Calabria does not necessarily imply to speak the local dialect; {\bf E1 + E4} together represent a severe and precise evidence, since the time slots when the criminal with red jacket and Valjean were seen riding the same scooter were too close to allow a change in the scooter's driver. Finally, {\bf E5} is both severe and precise, as it proves that Valjean was in the crime scene at the time when the crime was perpetrated.\\

\noindent The correct answers for the questions above are hence the following:

\noindent -- {\bf A1}, false: the availability of {\bf E1, E2, E3} represent the real setting of the case of B.C., and indeed B.C. was acquitted by the judge Giuseppe Gennari on June 18th, 2015.

\noindent -- {\bf A2}, true: {\bf E1 + E4} is severe and precise and it is coherently supported by {\bf E2} and {\bf E3}. 

\noindent -- {\bf A3}, false: if Thenardier is unreliable, {\bf E4} cannot be considered any longer and the situation becomes the same as in {\bf A1}, where the only evidences that could be used were {\bf E1, E2, E3}. 

\noindent -- {\bf A4}, true: {\bf E5} is severe and precise and it is coherently supported by the other evidences.

\section{The Logical Judge and the Case of Valjean}
\label{section:logicaljudge}

\begin{figure}[!ht]
    \centering
    \includegraphics[width=1\linewidth]{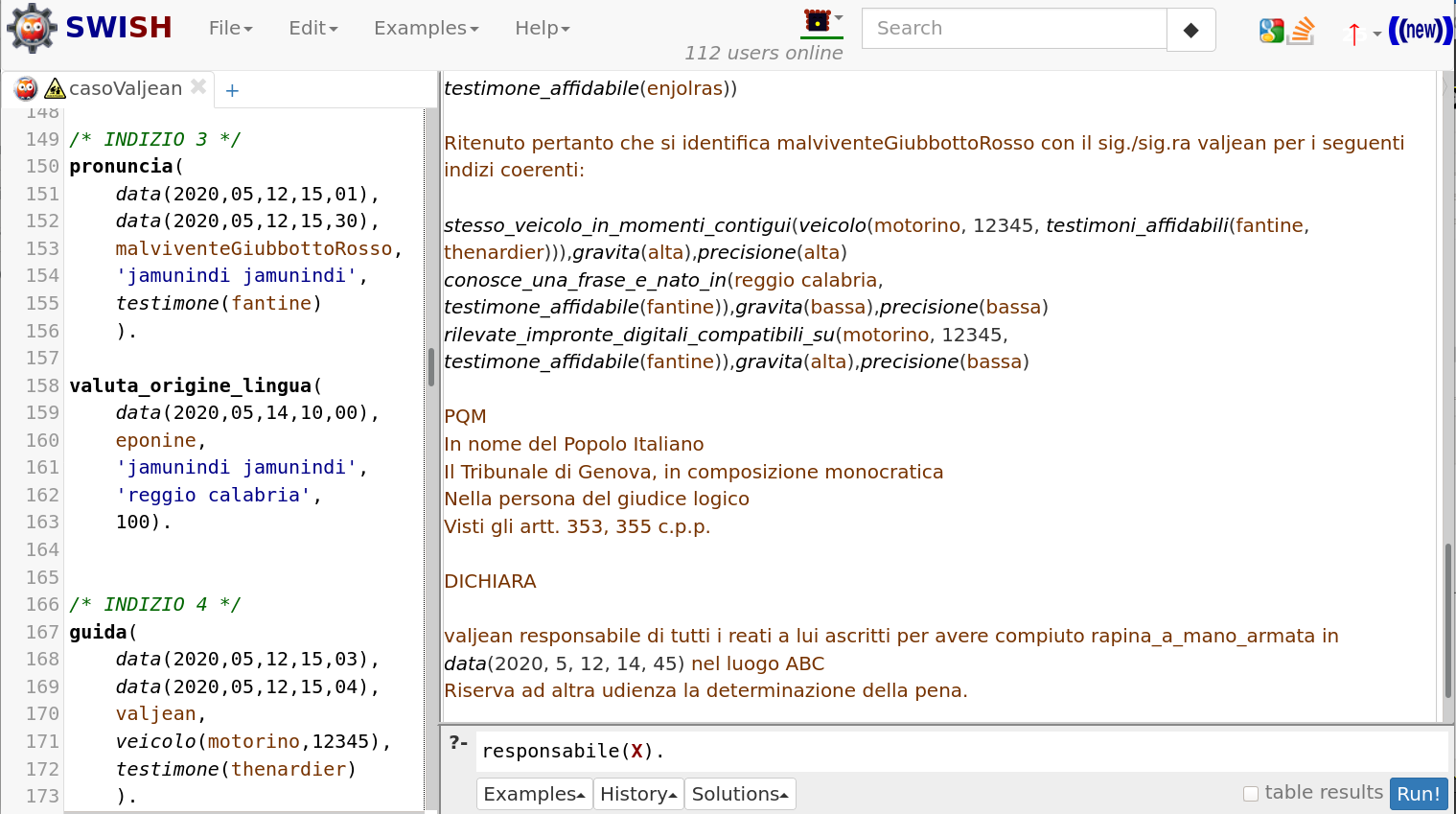}
    \caption{Screenshot of the logical judge on SWISH.}
    \label{fig:swish}
\end{figure}

The B.C. case offuscated by using different names and the code of the logical judge are implemented in SWISH\footnote{SWISH is a web front-end for SWIProlog available from \url{https://swish.swi-prolog.org}, last accessed on June 2020.}  and can be accessed at \url{https://swish.swi-prolog.org/p/casoValjean.pl}.
Figure \ref{fig:swish} shows a screenshot from that web page. 

All the facts and evidences presented in Section \ref{section:valjeanjavert} have been translated into Prolog facts, where predicates, functors, constants and variables are Italian words to allow the audience of the SSM course to immediately grasp their meaning. As an example, Figure \ref{fig:swish}, left pane, shows three facts that could be translated in English as\\

{\footnotesize{
\begin{tabularx}{\linewidth}{ X X X}
    \begin{minipage}[t]{0.29\textwidth}
   {\texttt{
\noindent {\bf /* EVIDENCE 3 */} \\
utters(\\
\hspace*{.2cm}    date(2020,05,12,15,01),\\ 
\hspace*{.2cm}    date(2020,05,12,15,30), \\
\hspace*{.2cm}    criminalInRedJacket, \\
\hspace*{.2cm}    'jamunindi jamunindi', \\
\hspace*{.2cm}    witness(fantine)).
}}                   
                \end{minipage}  
&
    \begin{minipage}[t]{0.38\textwidth}
\vspace*{.22cm}
{\texttt{
\noindent words\_origin\_evaluation(\\
\hspace*{.2cm}    date(2020,05,14,10,00),\\ 
\hspace*{.2cm}    eponine, \\
\hspace*{.2cm}    'jamunindi jamunindi',\\ 
\hspace*{.2cm}    'reggio calabria', \\
\hspace*{.2cm}    100).
    }}
     \end{minipage}
     &
    \begin{minipage}[t]{0.31\textwidth}
 {\texttt{
\noindent {\bf /* EVIDENCE 4 */}\\
drives(\\
\hspace*{.2cm}    date(2020,05,12,15,03),\\
\hspace*{.2cm}    date(2020,05,12,15,04),\\
\hspace*{.2cm}    valjean, \\
\hspace*{.2cm}    vehicle(scooter,12345), \\
\hspace*{.2cm}    witness(thenardier)).}}
     \end{minipage}
\end{tabularx}
}}

\noindent Information on Valjean, the witnesses, and the armed robbery are also modelled by Prolog facts like\\

{\footnotesize{
\begin{tabularx}{\linewidth}{ X X X}
    \begin{minipage}[t]{0.28\textwidth}
   {\texttt{
\noindent  born(\\
\hspace*{.2cm}     date(1980,10,17,13,07),\\ 
\hspace*{.2cm}     valjean, \\
\hspace*{.2cm}     'reggio calabria'). }}
                \end{minipage}  
&
    \begin{minipage}[t]{0.26\textwidth}
{\texttt{
\noindent commits( \\
\hspace*{.2cm}     date(2020,05,12,14,45),\\ 
\hspace*{.2cm}     criminalInRedJacket, \\
\hspace*{.2cm}     armedRobbery, \\
\hspace*{.2cm}     witness(enjolras)).
    }}
     \end{minipage}
     &
    \begin{minipage}[t]{0.44\textwidth}
\noindent {\texttt{
\hspace*{-.4cm} reliable(enjolras, hi).\\
reliable(fantine, hi). \\
reliable(thenardier, hi). 
}}
     \end{minipage}
\end{tabularx}
}}
\noindent The reasoning mechanism is based on defining under which conditions one evidence backs up the fact that two individuals are the same one (for example, the fact that \texttt{X} and \texttt{Y}, with \texttt{X} $\setminus =$ \texttt{Y}, were seen driving the same scooter in very close instants by two reliable witnesses is a highly severe and precise evidence that \texttt{X} and \texttt{Y} are same person), and collecting all those evidences, whose precision and severity may differ. If the evidences are at least two, supported by reliable witnesses, and at least one of them is severe and precise, we deduce that the two individuals are the same thanks to the following rule:
\\
{\footnotesize{
{\texttt{
\noindent same\_person(X, Y, Evidences) :- \\
\hspace*{.2cm}  setof((Ev, severity(S), precision(P)), \\
\hspace*{1.5cm}       evidence\_same\_as(Ev, X, Y, severity(G), precision(P)),\\ 
\hspace*{1.5cm}       Evidences),\\
\hspace*{.2cm}  length(Evidences, L),  L $>$ 1, member((\_, severity(hi), precision(hi)), Evidences).
  }}}}
\\

\noindent The definition of \texttt{responsible(X)} is the following
\\
{\footnotesize{
{\texttt{
\noindent responsibile(X) :- \\
\hspace*{.2cm}   committed(Y, Date, Crime, Place, EvidCrimeCommitted),   \\
\hspace*{.2cm}   same\_person(X, Y, EvidSamePerson), \\
\hspace*{.2cm}   pretty\_print(Date, X, Y, Crime, Place, EvidCrimeCommitted, EvidSamePerson).
}}}}   
\\

\noindent where \texttt{committed} is defined by exploiting evidences and witnesses that support the fact that \texttt{Y} committed a crime and \texttt{pretty\_print} prints the text shown in Figure \ref{fig:swish}, right pane. The printed text includes the motivations for the trial outcome and the formula that Italian judges utter to state their final decision.  

By commenting and de-commenting evidences in the Prolog code and by changing the reliability of witnesses from \texttt{hi} to \texttt{lo}, we demonstrated to the audience that the different scenarios depicted in Section \ref{section:valjeanjavert} can be simulated, with the logical judge answering {\bf A1} to {\bf A4} in the correct way.  We also showed that changes to the Code of Criminal Procedure could be implemented by operating on the \texttt{same\_person} predicate. For example, we might change \texttt{L} $> 1$ into \texttt{L} $> 0$ and state that only one evidence is enough, provided that it is precise and severe, which is actually the case of the Italian law. Or we might change it into \texttt{L} $> 3$ and remove \texttt{member((\_, severity(hi), precision(hi)), Evidences)}, stating that at least four coherent evidences are considered as a proof, even if none is severe and precise. 
\\

\begin{figure}[!ht]
    \centering
    \includegraphics[width=1\linewidth]{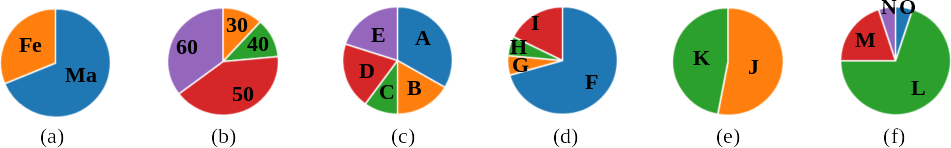}
    \caption{Figures from the post-course survey.}
    \label{fig:survey}
\end{figure}

%\section{The Human Judge}
%\label{section:feedbackattendees}
%
% \section{Discussion}
% \label{section:conclusions}
%
The questions and remarks we received during and immediately after the working group activities showed that the topic raised the interest of the audience, and we felt we had successfully achieved our dissemination goals. To measure this feeling in a scientific way, a few weeks after the working group we prepared a survey and invited the participants to answer a few questions.  
Figure \ref{fig:survey} shows the results of this survey, aimed at assessing the background of the attendees and getting their feedback on the logical judge. Only 17 attendees answered the questionnaire, which was not mandatory and anonymous.
%out of 70 and more answered the questionnaire, which was not mandatory, anonymous, and carried out via a GDPR-compliant tool.  

\noindent -- Chart {\bf (a)} shows the gender of those who participated in the survey: 11 males (69\%) and 5 female (31\%) (one participant did not answer). 

\noindent -- Chart {\bf (b)} shows their age: 2 between 30 and 39 year old (12\%), 2 between 40 and 49 (12\%), 7 between 50 and 59 (41\%), and 6 between 60 and 69 (35\%).

\noindent -- Chart {\bf (c)} summarizes the participants' background on Artificial Intelligence, before following the SSM course; multiple answers to the questions about synonymy of popular locutions were allowed. 10 participants were aware that Artificial Intelligence and Algorithm are not synonym (slice A of the pie); 5 were aware that AI and Machine Learning are not (slice B); 3 knew that  AI and Natural Language Processing are not (slice C); 6 knew that  AI and Automated Reasoning are not (slice D). 6 participants stated that none of the above assertions was true for them (slice E). This last result is not surprising given the confusion about AI and the misuse of the term in non-technical (sometimes, also in technical) articles, but it is nevertheless worth some serious consideration that 6 participants out of 17 were not sure about the real meaning of popular words and terms like AI, algorithm, machine learning.

\noindent -- Chart {\bf (d)} presents the answers to our question on previous knowledge about logic programming languages. 12 participants did not know about their existence, but they appreciated their potential (71\%, slice F), one was not aware of their existence and did not appreciate it, since  he/she saw no applicability for them (6\% , slice G); one was already aware of their existence (6\%, slice H). 3 answered ``Other''.
%(17\%, slice H).

\noindent -- Chart {\bf (e)} summarizes the answers about the possibility to have logical judges substituting, up to some extent, human judges. 9 participants felt that such approaches could offer a significant support to the judge in the future, but will never substitute her (53\%, slice J), and 8 participants felt that the support given by such kind of approaches will be limited in the future (47\%, slice K).

\noindent -- Chart {\bf (f)} presents the main feelings raised by logical judges. Multiple answers were allowed: 14 participants stated to be curious about their potential and limitations (slice L); 4 were looking for concrete results (slice M);  one asserted to be worried about their potential raise (slice O). One answer was ``Other''. \\

Charts {\bf (e)} and {\bf (f)} support our feeling that we were able to raise the curiosity in the audience without neither scaring them, nor creating false expectations. We stress that our activities were carried out in less than two hours, and the background of the attendees was almost basic, as shown by Charts {\bf (c)} and {\bf (d)}. One comment that emerged many times, both during the working group and as a free comments in the survey, is the need for judges to be exposed to a scientifically grounded and honest introduction to AI, in order to become informed and active players and decision-makers of their own future. \\

 \noindent {\bf Acknowledgements.} This paper is based upon work from COST Action DigForASP, supported by COST (European Cooperation in Science and Technology). 
 %We acknowledge all the DigForASP partners for the exciting and constructive discussions.

%\nocite{*}

\end{document}